\documentclass[twoside]{article}
\usepackage[accepted]{style/aistats2019}

\usepackage[round]{natbib}

\setcitestyle{authoryear,open={(},close={)}}
\newcommand{\citeay}[1]{\citeauthor{#1}, \citeyear{#1}}

\usepackage{microtype}
\usepackage{bibentry}

\usepackage{marginnote}  

\usepackage{stfloats}

\usepackage{graphicx} 
\usepackage{subcaption} 

\usepackage{bm}

\usepackage{amsmath}
\usepackage{amssymb}

\usepackage{commath} 

\usepackage[normalem]{ulem}

\usepackage{soul}

\usepackage{numprint}

\usepackage{tikz}
\usetikzlibrary{arrows, arrows.meta, shapes.geometric}

\usepackage{scrextend}

\usepackage{newfloat}
\usepackage{caption}

\usepackage{booktabs}

\usepackage{newfloat}
\usepackage{caption}
\DeclareFloatingEnvironment[fileext=frm,placement={tb},name=Frame]{myfloat}
\usepackage[framemethod=1,leftmargin=0pt,rightmargin=0pt]
{mdframed} 


\usepackage[hidelinks]{hyperref}

\usepackage{multicol}

\usepackage{ifthen}
\newboolean{showControlInFig}

\usepackage{expl3, xparse}
\ExplSyntaxOn
\keys_define:nn { mbfinternal / mbf } {
  latin          .tl_set:N = \mbfinternal_mbf_latin_tl,
  greek          .tl_set:N = \mbfinternal_mbf_greek_tl,
  latin-alphabet .tl_set:N = \mbfinternal_mbf_latin_alphabet_tl,
  greek-alphabet .tl_set:N = \mbfinternal_mbf_greek_alphabet_tl,
}
\cs_new:Nn \mbfinternal_mbf:n {
  \tl_if_in:NnTF \mbfinternal_mbf_latin_alphabet_tl { #1 } {
    \mbfinternal_mbf_latin_tl { #1 }
  } {
    \tl_if_in:NnT \mbfinternal_mbf_greek_alphabet_tl { #1 } {
      \mbfinternal_mbf_greek_tl { #1 }
    }
  }
}
\NewDocumentCommand \mbf { O{} m } {
  \group_begin:
  \keys_set:nn { mbfinternal / mbf } {
    latin-alphabet = abcdefghijklmnopqrstuvwxyz
                     ABCDEFGHIJKLMNOPQRSTUVWXYZ
                     0123456789,
    greek-alphabet = \alpha\beta\delta\epsilon
                     \phi\gamma\eta\iota\theta
                     \kappa\lambda\mu\nu\pi\chi
                     \rho\sigma\tau\omega\xi\psi\xi
                     \Alpha\Beta\Delta\Epsilon  
                     \Phi\Gamma\Eta\Iota\Theta  
                     \Kappa\Lambda\Mu\Nu\Pi\Chi 
                     \Rho\Sigma\Tau\Omega\Xi\Psi\Xi, 
    latin = \mathbf,
    greek = \boldsymbol,
    #1
  }
  \mbfinternal_mbf:n { #2 }
  \group_end:
}
\ExplSyntaxOff

\newcommand{\alexnote[1]}{[\textit{\textcolor{red}{#1}}]}

\newcommand*{\Tr}{^{\mkern-1.5mu\mathsf{T}}}

\usepackage{array}

\newboolean{isanonymised}
\setboolean{isanonymised}{True}

\newcommand{\anonymousswitch}[2]{\ifthenelse{\boolean{isanonymised}}{\alexnote[#1]}{\textcolor{orange}{#2 [-- for anon]}}}
\newcommand{\anonymousswitchnoformat}[2]{\ifthenelse{\boolean{isanonymised}}{#1}{#2}}

\newboolean{bshownotes}
\setboolean{bshownotes}{False}
\newcommand{\showalexnotes}[1]{\ifthenelse{\boolean{bshownotes}}{#1}{}}

\makeatletter
\newcommand*{\cleartoleftpage}{%
  \clearpage
    \if@twoside
    \ifodd\c@page
      \hbox{}\newpage
      \if@twocolumn
        \hbox{}\newpage
      \fi
    \fi
  \fi
}
\makeatother

\begin{document} 
\twocolumn[

\aistatstitle{Multi-Task Time Series Analysis applied to Drug Response Modelling}

\aistatsauthor{ Alex Bird \And Christopher K. I. Williams \And  Christopher Hawthorne }

\aistatsaddress{ 
\begin{minipage}{0.3\textwidth}\centering 
School of Informatics,\\ University of Edinburgh, UK \\
The Alan Turing Institute, UK
\end{minipage}
\And  \begin{minipage}{0.3\textwidth}\centering School of Informatics,\\ University of Edinburgh, UK \\ The Alan Turing Institute, UK \end{minipage} \And 
\begin{minipage}{0.3\textwidth} \centering
NHS Greater Glasgow \\ and Clyde, UK \\
\vphantom{padding}
\end{minipage}
} 

]

\setlength{\marginparwidth}{2cm}
\newcommand{\Blue}{\textcolor{blue}}
\newcommand{\Green}{\textcolor{red}}
\newcommand{\fix}{\marginpar{FIX}}
\newcommand{\switchup}{black}

\begin{abstract}
Time series models such as dynamical systems are frequently fitted to
a cohort of data, ignoring variation between individual entities such
as patients. In
this paper we show how these models can be personalised to an
individual level while retaining statistical power, via use
of \emph{multi-task learning} (MTL). To our knowledge this is a novel
development of MTL which applies to time series both with and without
control inputs. The modelling framework is demonstrated on a
physiological drug response problem which results in improved
predictive accuracy and uncertainty estimation over existing
state-of-the-art models.
\end{abstract}


\section{INTRODUCTION}
\label{sec:introduction}
In this paper we study the response of a patient's physiology under the effect
of drug infusion, specifically the effect of the anaesthetic  agent Propofol
on vital signs. There is a long history of pharmacokinetic/pharmacodynamic
(PK/PD) models used to study such effects (see e.g.\
\citeay{bailey2005drug}, and
\citeay{minto2008contributions}). We
are particularly interested in the \emph{personalisation} of such models to
individual patients. Our focus here will be the on the personalisation of the
pharmacodynamic (PD) model, as this issue has already been addressed for the
PK component (see e.g.\ \citeay{eleveld2018pharmacokinetic}).

We approach this problem as one of \emph{multi-task learning} (MTL),
where each patient is treated as a task. We show that the parameters
across patients can be well-modelled as a low-dimensional latent
linear subspace.  Our results indicate that approximately five such
latent variables suffice to describe the inter-patient variation in
response
we observe in a clinical study, and lead to
improved predictive accuracy and uncertainty estimation over existing
state-of-the-art models. 

The application of MTL to time series models is unusual, especially for individuals unrelated in space and time. In this work we generalise multi-task learning to tasks described by dynamical systems -- an important class of time series models. 
Comparison to existing work is given in Section \ref{sec:relatedwork} including the new classes of models to which we extend the MTL framework.


The structure of the paper is as follows:  Section \ref{sec:pkpd} provides an overview of the PK/PD model and its personalisation. Section \ref{sec:model} introduces our proposed model; related work is discussed in Section \ref{sec:relatedwork}. Our results on clinical data are summarised and discussed in Section \ref{sec:experiments}, and some concluding remarks are given in Section \ref{sec:conclusion}.

\section{PHARMACOKINETIC-PHARMACODYNAMIC (PK/PD) MODELS}
\label{sec:pkpd}
PK/PD models are the dominant paradigm for modelling the response to a
continuously infused drug. We use the anaesthetic agent Propofol as our running example. These models decompose the problem into two sub-tasks. Of primary interest in the literature is the \emph{pharmacokinetic} (PK) process\footnote{The origin of this model can be traced back at least as far as \citet{jelliffe1967mathematical}.}, which relates the administration of a drug to its distribution and elimination in the body over time. This provides insight into the evolution of drug concentrations in regions of the body including major blood vessels. A \emph{pharmacodynamic} (PD) process models the relationship between blood concentration and the effect on observed physiological effects, such as vital signs (e.g.\ heart rate, blood pressure).
Below, we briefly discuss common approaches to these models.

\paragraph{PK models.} The most commonly described PK model in anaesthesia is the three-compartment model. This is an Ordinary Differential Equation (ODE) describing drug concentrations for differently perfused physiological compartments under a continuous time infusion. These compartments notionally correspond to blood, muscle and fat, and can be conceptualised with the latter two as `peripheral' compartments each connected only to the \emph{central compartment}, blood. 
The drug in-flow from a Target Controlled Infusion (TCI) pump, $\omega(t)$ enters the central compartment and the concentrations $\mbf c(t) = (c_1(t),\, c_2(t),\, c_3(t))\Tr$ of all compartments evolve as:
\begin{align}
    \frac{\dif \mbf c}{\dif t} = A \mbf c(t) + \mbf e_1  \omega(t)&  \label{eq:PK}
\end{align}
for a given matrix $A$ of rate constants (see supplementary material), and the first ordinate unit vector $\mbf e_1$. There is a large body of literature pertaining to the PK model and it exhibits strong performance in experimental and clinical settings. Much work has been done to personalise the rate constants in the matrix $A$ based on patient covariates (see below). It is notable that the rate constants have become part of the vocabulary of practicing anaesthetists.

\paragraph{PD models.} The mechanism by which physiological effects follow from the blood concentration of some drug is not well understood; PD models propose a direct relationship to the concentration at some physiological \emph{effect site}. The effect site concentration $x(t)$ may not have reached equilibrium with the central compartment concentration $c_1(t)$ and can introduce a lag to the observed effect. Denoting the rate constants as $k_{1e}$ for the in-flow to the effect site and $k_{e0}$ for the elimination, the relationship is modelled as: 
\begin{align}
    \frac{\dif x}{\dif t} = k_{1e} c_1(t) - k_{e0} x(t). \label{eq:PD-ODE}
\end{align} 
For multidimensional observations, independent effect sites $x_j(t), j = 1,\ldots d$ are usually fitted. For instance, the observation channels in our dataset consist of systolic and diastolic blood pressure (BPsys and BPdia respectively), and the Bispectral Index\footnote{The Bispectral Index of \citet{myles2004bispectral} is a proprietary scalar-valued transformation of EEG signals which attempts to quantify the level of consciousness.} 
(BIS). The relationship of the effect site to observations $y_{j}(t), j = 1,\ldots d$ are then modelled by some nonlinear transformation plus white noise, i.e.\ for a given time $t$, $y_{j}(t) = \mathcal{N}(g_{\theta}(x_j(t)), \tau^{-1})$. Common choices of $g_{\theta}(\cdot)$ are the Hill function \citep{hill1910possible} or generalised logistic sigmoid (see e.g.\ \citealt{georgatzis2016ionlds}):
\begin{align}
    g_\theta(x) = \theta_1 + \frac{\theta_2 - \theta_1}{(1 + \exp\{-\theta_3 x\})^{1/\theta_4}}.
\end{align}

\paragraph{Personalisation.} A number of attempts have been made to personalise PK models using patient attributes such as age, gender, height or weight (see e.g.\ \citealt{marsh1991pharmacokinetic, schnider1998influence, white2008use} and \citeay{eleveld2018pharmacokinetic} for a combined study). Various studies \citep[see e.g.][]{glen2009evaluation, masui2010performance, glen2014comparison} have compared the predictive performance of several propofol PK models currently used for target controlled infusion (TCI) in clinical practice. These studies have confirmed a degree of bias and inaccuracy of the models but overall their performance is considered by most clinicians to be adequate for clinical use (at least within the populations in which they were developed).

In most commercially available implementations of the Marsh and Schnider models, fixed $k_{1e}, k_{e0}$ are used, as well as constant parameters in the emission function $g_\theta(\cdot)$. This means that there is no adjustment of the PD component of the model based on patient covariates. It is widely accepted by practicing anaesthetists that there is a significant amount of inter-individual variability in PD response to Propofol. As it is the clinical effect, rather than drug concentration, that is most important in clinical practice, we therefore focus on improving the PD component of the model. Other work has investigated re-fitting the model end-to-end \citep{georgatzis2016ionlds} but given the quantities of data available, we believe better results are available by leaving the PK model unaltered, as well as retaining better interpretability.

\section{PROPOSED MODEL}
\label{sec:model} Our proposed model is a discrete time model closely following the existing framework for PD models in Section \ref{sec:model:singletask}. This model can be fitted individually to patients (a `single-task' approach), but many observations are required before useful predictions can be made. Section \ref{sec:model:multitask} then introduces a multi-task variant which meets our criterion of personalisation while reducing the sample complexity. Implementation is discussed in the final section.

\subsection{Parameterisation of PD model} \label{sec:model:singletask}
Denote the central compartment solution, $c_1(t)$, to the PK model in eq.\  (\ref{eq:PK}) over a uniform time grid as $\{u_t\}_{t=1}^T$ which we use as the input for our PD model.  
The observations, $\{\mbf y_t\}_{t=1}^T$ with $\mbf y_t \in \mathbb{R}^d$ are modelled with independent effect sites for each each observation channel. We denote the concentration at each effect site $j$ by $\{x_{tj}\}_{t=1}^T$, and for this time grid, the effect site relationship (\ref{eq:PD-ODE}) may be written:
\begin{align}
    x_{tj} &= \beta_{j1} u_{t-1} + \beta_{j2} x_{t-1,j}  \label{eq:vanilla-pd-recurrence}
\end{align}
with no loss of generality if $c_1(t)$ is piecewise constant.

In this case application of the convolution theorem gives $\beta_{1} = \frac{k_{1e}}{k_{e0}} (1 - e^{-k_{e0}})$ and $\beta_{2} = e^{-k_{e0}}$ (omitting channel index $j$ for clarity, see supplementary material). Since both rate constants are positive, this implies for all $j$, $\beta_{j1} > 0$ and $\beta_{j2} \in (0,1)$, with the latter enforcing stability and non-oscillation of the AR process in (\ref{eq:vanilla-pd-recurrence}). 

For the nonlinear emission, we require a parametric function $g_{\theta}(\cdot)$ for which previous choices (as in Section \ref{sec:pkpd}) have proved to be numerically unstable or insufficiently expressive. We instead use a basis of $L$ logistic sigmoid functions, $\sigma(x) = \frac{1}{1 + \exp\{-x\}}$, and express $g_\theta(x) = \sum_{r=1}^L \theta_r\, \sigma(\,a_r (x - b_r))$ with slopes $\{a_r\}$ and offsets $\{b_r\}$. With appropriate optimisation of the $\{a_r, b_r\}$, we found that $L=8$ basis functions sufficed to well approximate the generalised sigmoid as used in \citet{georgatzis2016ionlds}. The constraints $a_r < 0$ and $\theta_r \ge 0$ for all $r$ ensures the monotonic property that as concentration increases, the observations are non-increasing.

The full PD model can now be written for each patient $i \in 1, \ldots N$
(denoted by a superscript) as:
\begin{subequations}
\label{eq:PD-indiv}
\begin{align}
    \text{\footnotesize (dynamics)}\quad x_{tj}^i &= \beta_{j1}^i u_t^i + \beta_{j2}^i x_{t-1,j}^i \label{eq:PD-indiv-recurrent} \\
    \text{\footnotesize (pre-activation)}\quad \mbf h_{tj}^{i} &=  (x_{tj}^{i}\mbf 1 + \beta_{j3}^{i}\mbf 1 - \mbf b) \circ \mbf a \label{eq:PD-indiv-hidden}\\
    \text{\footnotesize (emission)}\quad y_{tj}^{i} &= \sigma(\mbf h_{tj}^{i})\Tr \mbf \theta_j^{i} + \alpha_j^{i} + \epsilon_{tj}^{i} \label{eq:PD-indiv-emission}
\end{align}
\end{subequations}
for $\epsilon_{tj}^{i} \sim \mathcal{N}(0, \tau^{-1}), \, j \in 1, \ldots d_i \text{
and } t \in 1, \ldots T_i$. Element-wise multiplication is denoted by $\circ$, and $\sigma(\cdot)$ is overloaded to act elementwise on multivariate inputs. In order to permit greater modelling flexibility we also introduce parameters $\mbf \beta_3$ and $\mbf \alpha$ which provide a personalised offset to the values of the effect site dynamics and the emission respectively. The learnable parameters are now $\tau$ and $\{\mbf \alpha^i, \mbf \beta_1^i, \ldots, \mbf \beta_d^i, \mbf \theta_1^i, \ldots, \mbf \theta_d^i\}$, $i \in 1, \ldots, N$. The $\alpha_j^{i}$ relate to pre-infusion patient specific vitals levels which could be estimated in advance of anaesthetic induction if data are available.

\subsection{Proposed multi-task model} \label{sec:model:multitask}
By considering each patient $i$ as a task, we can use a low-rank multi-task structure to share information between patients. Define a latent variable $\mbf z^i \in \mathbb{R}^k$ as the low dimensional representation of the task parameters, and a loading matrix $\Psi \in \mathbb{R}^{p\times k}$ in analogy with Factor Analysis. We can relate this to the individual task parameters via the following model:
\begin{subequations}
\label{eq:gen-model}
\begin{align}
    \mbf z^i\,\, &\sim \,\,\mathcal{N}(0, I)\\
    \mbf \lambda^i\,\, &= \,\, \mbf f(\Psi \mbf z^i)\\
    \mbf \beta_{1}^i,\: \mbf \beta_{2}^i,\: \mbf \beta_{3}^i,\: \mbf \theta_{1}^i,\: \ldots, \mbf \theta_{d}^i,\: \mbf \alpha^i &=\,\, \operatorname{unpack}(\mbf \lambda^i) \qquad\label{eq:MTL-unpack}
\end{align}
\end{subequations}
with each of the parameters in (\ref{eq:MTL-unpack}) serving as the multi-task parameters in the deterministic state space model (\ref{eq:PD-indiv}). $\mbf f $ is a pre-specified vector-valued function acting componentwise on $\Psi \mbf z$ in a potentially nonlinear way (see below). The `unpack' operation here partitions the vector $\mbf \lambda^i$ into dimensionally consistent quantities as implied by the LHS. Assuming a Bayesian approach to fitting (\ref{eq:PD-indiv}), this results in a reduction of latent variables from $Np$ to $Nk$ vs.\ a single task approach.

The function $\mbf f$ is introduced primarily in order to satisfy the constraints required for (\ref{eq:PD-indiv}) in Section \ref{sec:model:singletask}, and is defined elementwise by various univariate transformations. For example, the  non-negativity constraints for $\{\mbf \beta_1^i\}$ can be enforced by $\text{softplus}(x) = \log(1 + e^x)$ and the unit interval for $\{\mbf \beta_2^i\}$ by a logistic sigmoid, etc. For all unconstrained parameters, such as offsets, no nonlinearity is applied. (If desired, more elaborate transformations may be considered.) The use of an elementwise $\mbf f$ results in any multivariate relationships to be approximated by the covariance structure learned in $\Psi$. \begin{figure}[tb!]
\centering
\setlength{\marginparwidth}{1.6cm}
\setlength{\marginparsep}{1mm}
\setlength{\marginparwidth}{2.0cm}
\setlength{\marginparsep}{3mm}
\scalebox{0.60}{
\begin{tikzpicture}[latentcts/.style={circle, minimum size=31pt,draw},
                    latentdiscrete/.style={minimum size=31pt,draw},
                   latentdeterministic/.style={diamond, minimum size=37pt,draw,inner sep=0pt},
                    visiblects/.style={circle,minimum size=31pt,fill=purplevis!50, draw},
                    visiblediscrete/.style={minimum size=31pt,fill=purplevis!50, draw,},
                    invis/.style={minimum size=0pt,fill=white, inner sep=0pt},
                    dot/.style 2 args={circle,inner sep=1pt,fill,label={#2:#1},name=#1},]
\definecolor{purplevis}{RGB}{185,164,219}
\definecolor{bluedeterministic1}{RGB}{75, 136, 210}
\definecolor{bluedeterministic2}{RGB}{158, 192, 235}
\definecolor{bluedeterministic1}{RGB}{0, 0, 0}
\definecolor{bluedeterministic2}{RGB}{130, 130, 130}

\def\numseq{4}
\def\numvars{2}  

\pgfmathtruncatemacro{\numseqm}{\numseq - 1}

\draw (-1,-1.1) -- (10.7,-1) -- (10.7,5.7) -- (-1,5.7) -- (-1,-1.1);

  \foreach \x in {0,...,\numseq}            
    \foreach \y in {0,...,\numvars} 
       {\pgfmathtruncatemacro{\label}{\x + 1}
       \ifnum \y = 0
            \def\varlbl{y}
            \def\tzndstyle{visiblects}
        \else
            \ifnum \y = 1
                \def\varlbl{x}
                \def\tzndstyle{latentdeterministic}
            \else  
                \def\varlbl{u}
                \def\tzndstyle{}
            \fi
        \fi
        \def\tzndlbl{$\mathbf{\varlbl}_{\label}^{i}$}
        \def\tzndxpos{2.5*\x + 0.25}
        
        \ifnum \x = \numseq
            \def\tzndlbl{}
            \def\tzndstyle{invis}
            \def\tzndxpos{2.5*\x - 0.5}
        \fi
        
       \node [\tzndstyle]  (\y\x) at (\tzndxpos, 2*\y) {\Large \tzndlbl}
       ;} 

  \foreach \y  in {1,...,1}
    \node[anchor=center] at (2.5*\numseq, 2*\y) {$\ldots$};

\def\zposx{3.8}
\def\zposy{4.8}

\node [latentcts]  (zi) at (\zposx, \zposy) {\Large $\mbf z^{i}$};
\draw[-{Latex[length=2mm]}, bluedeterministic2] (zi) to [out=180,in=70] (10);
\draw[-{Latex[length=2mm]}, bluedeterministic2] (zi) to [out=230,in=70] (11);
\draw[-{Latex[length=2mm]}, bluedeterministic2] (zi) to [out=195,in=65] (00);
\draw[-{Latex[length=2mm]}, bluedeterministic2] (zi) to [out=265,in=60] (01);
\draw[-{Latex[length=2mm]}, bluedeterministic2] (zi) to [out=280,in=120] (02);
\draw[-{Latex[length=2mm]}, bluedeterministic2] (zi) to [out=345,in=115] (03);
\draw[-{Latex[length=2mm]}, bluedeterministic2] (zi) to [out=310,in=110] (12);
\draw[-{Latex[length=2mm]}, bluedeterministic2] (zi) to [out=0,in=110] (13);

\pgfmathsetmacro{\zposxm}{\zposx - 0.7}%
\pgfmathsetmacro{\zposxp}{\zposx + 1.2}%
\pgfmathsetmacro{\zposyp}{\zposy + 1.7}%

  \foreach \y in {1,...,1}  
    \foreach \x in {0,...,\numseqm}
      \pgfmathtruncatemacro{\xp}{\x + 1}%
      \draw[-{Latex[length=2mm]}, bluedeterministic1] (\y\x)--(\y\xp)
      ;
      
  \def\y{1}
  \foreach \x in {0,...,\numseqm}
    \pgfmathtruncatemacro{\ym}{\y - 1}%
    \draw[-{Latex[length=2mm]}, black] (\y\x)--(\ym\x)
  ;
  \def\y{2}
  \foreach \x in {0,...,\numseqm}
    \pgfmathtruncatemacro{\ym}{\y - 1}%
    \draw[-{Latex[length=2mm]}, bluedeterministic1] (\y\x)--(\ym\x)
  ;

\node [invis] (platei) at (9.5,-0.6) {$i=1,\ldots, N$};
\end{tikzpicture}
}
\caption{Multi-task input-output dynamical system. 
Stochastic nodes are denoted by circles and deterministic nodes by diamonds; arrows indicate functional relationships.}

\label{fig:mtl-pd}
\end{figure}
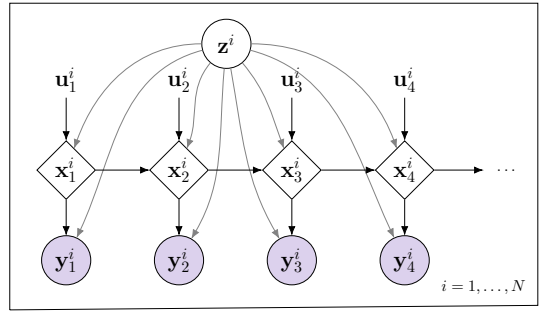

Collecting all task observations as $Y = \{\mbf y_{1:T_i}^i\}_{i=1}^N$ and similarly $U = \{ u_{1:T_i}^i\}_{i=1}^N$, $Z = \{\mbf z^i\}_{i=1}^N$, the joint distribution over observations and latent variables is:
\begin{align}
    p(Y, Z \mid U ; \Psi) &= \prod_{i=1}^N p(\mbf
    y_{1:T_i}^i \mid \mbf z^i, u_{1:T_i}^i; \Psi)\, p(\mbf z^i) . \label{eq:joint-probabilistic-model}
\end{align}
See Figure \ref{fig:mtl-pd} for a graphical model. While the likelihood is time structured, observations are conditionally independent given $\mbf z^i$ due to the deterministic evolution. The resulting decomposition yields a nonlinear non-\emph{iid} hierarchical model: $p(\mbf y_{1:T_i}^i \mid \mbf z^i, u_{1:T_i}^i; \Psi) = \prod_t p(\mbf y_{t}^i \mid \mbf z^i, u_{1:t}^i; \Psi)$.

The above model has been formulated for the case that $\mbf z^i$ are unobserved, but some side information may be known about the task. For example, in our drug response problem, we may have patient covariates such as age, height, weight etc. which may (partially) describe the differences between tasks. In this case, the latent variables can be replaced by observed ones, which we call the \emph{task-descriptor} model. In the case that all dimensions of each $\{\mbf z^i\}$ are assumed known, test time predictions may be calculated for new tasks without need of an inference step. 

\subsubsection{Practical implementation}
\paragraph{Inference.} Inference of $Z$ in this model is challenging. For our experiments, we use a Monte Carlo approach in order to obtain an accurate representation of uncertainty. Unfortunately, while inference proceeds sequentially, Sequential Monte Carlo (SMC) methods will suffer from severe particle degeneracy due to the \emph{static model} formulation \citep{chopin2002sequential}. The proposed rejuvenation step in \citet{chopin2002sequential} is also quite expensive in our case. In the experimental results given, we use Hamiltonian Monte Carlo \citep[HMC, see e.g.][]{neal2011mcmc} using the implementation of NUTS \citep{hoffman2014no} in Stan \citep{carpenter2016stan}. Alternatively a more general SMC samplers approach \citep{del2006sequential} could have been used.

\paragraph{Learning.} Optimising the parameters $\Psi$ requires integrating eq.\ (\ref{eq:joint-probabilistic-model}) over $Z$. This can be approached iteratively via (Monte Carlo) Expectation Maximisation \citep[see e.g.][ch. 6.1]{mclachlan2007algorithm} or gradient methods using HMC, but is expensive; in practice we performed joint optimisation of the $Z$ and $\Psi$, i.e.\ $\arg\max_{Z, \Psi} p(Y, Z | U; \Psi)$ as an approximation of the objective. Learning can then proceed by use of gradient methods and automatic differentiation; our implementation used Adam \citep{kingma2014adam} using the PyTorch \citep{paszke2017automatic} framework.

\paragraph{Prediction.} The primary goal of our drug-dosing model is to provide improved predictions for anaesthetic induction. The predictive $r$-step ahead posterior $p(\mbf y_{t+1:t+r}^i| \mbf y_{1:t}^i; \Psi)$ is not available in closed form, but can be approximated using the Monte Carlo posterior of $Z$, $p(\mbf z^i| \mbf y_{1:t}^i; \Psi) \approx \frac{1}{M} \sum_{m=1}^M \delta(\mbf z^i - \mbf z^{i(m)})$ where $\delta(\cdot)$ is the Dirac delta function. Then $p(\mbf y_{t+1:t+r}^i| \mbf y_{1:t}^i; \Psi) =$
\begin{align}
     \hphantom{=} \int p(\mbf y_{t+1:t+r}^i| \mbf z^i; \Psi)\, p(\mbf z^i| \mbf y_{1:t}^i; \Psi) \dif \mbf z^i \nonumber \\
    \approx \,\frac{1}{M}\sum_{m=1}^M p(\mbf y_{t+1:t+r}^i| \mbf z^{i(m)}; \Psi)\,.\label{eq:posterior-predictive-mc}
\end{align}

\paragraph{Practical details.} We presume the existence of a training set such that $\Psi$ can be learned in an offline stage, and so (\ref{eq:posterior-predictive-mc}) only requires inference of $\mbf z^i$ at each relevant time point. In our experiments, taking 3000 samples using 2 chains (after a thin factor of 2) sufficed for suitable mixing and an effective sample size typically exceeding 100. In order to achieve good mixing, we found it helpful to initialise from the relevant MAP value, and where necessary, also supplied a custom mass matrix from preliminary runs. 

Our single task experiments (see below) also used HMC to infer the parameters, now in $p=36$ dimensional space. This problem is higher dimensional and also less constrained, and hence is much harder than the MTL experiments. We limited CPU time to $10\times$ that used by the MTL experiments, which we believe gives a fair comparison of the methods; in both cases we leave ourselves open to the possibility that some chains had not reached convergence.

Use of the MAP approximation for learning produces a scale degeneracy not present in the original formulation. For example shrinking a MAP estimate of $Z^{\text{MAP}}$ by a diagonal factor $D$ (with the inverse applied to $\Psi$) can result in an improved MAP objective value. This can be circumvented in practice by early stopping, and re-scaling to ensure $\operatorname{diag}(Z\Tr Z) = I$.

The low-rank structure confers benefits of both statistical strength and computational efficiency at test time, which is controlled by choice of $k$. While the effective dimensionality of the parameter variation is unknown, we choose $k$ based on model comparison (BIC) on preliminary experiments which suggested that $k=5, 7$ represented a good trade-off between individual and cohort performance.

\begin{figure*}[t!]
    \centering
    \includegraphics[trim={0 0 0 0}, clip, width=0.9\textwidth]{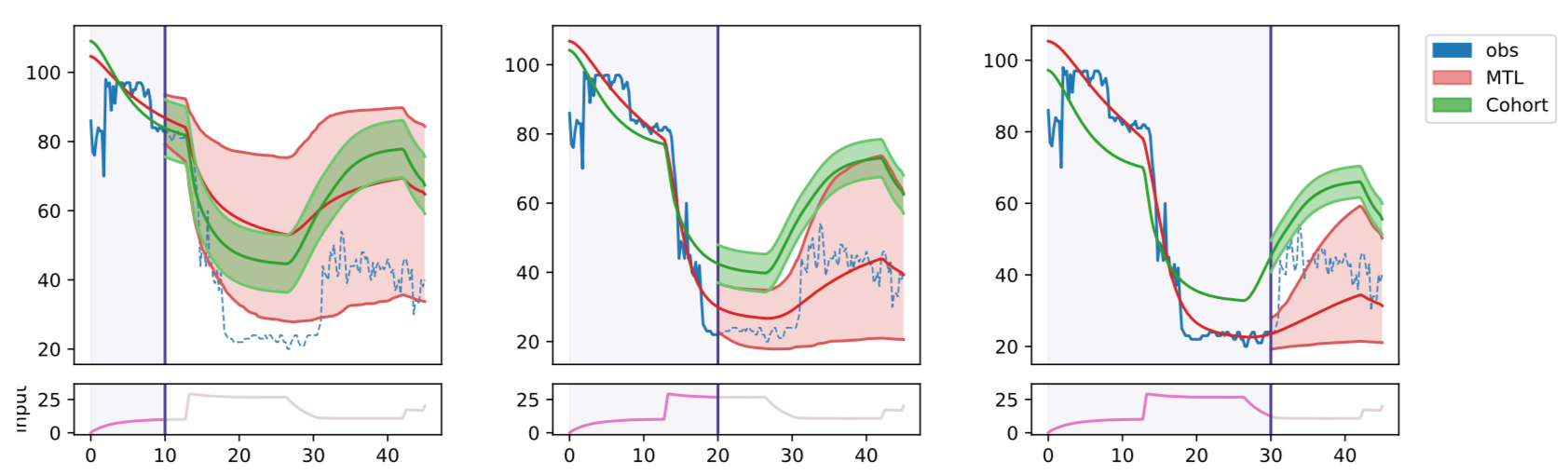}
    \caption{Example predictions for BIS channel at T=10, 20, 30 minutes (left-to-right) for cohort and MTL ($k=5$) models. Shown are the predictive mean and 90\% (approx.) credible interval of the underlying function. Retrospective fits are shown without intervals for clarity. Best viewed in colour.}
    \label{fig:BIS_example}
\end{figure*}

\section{RELATED WORK}
\label{sec:relatedwork}

There is a substantial literature on multi-task learning (MTL) when there is \emph{iid} input-output data for each task, see e.g.\ the review paper by \citet{zhang2017survey}. This may involve learning a hidden-unit representation shared across tasks \citep[see e.g.][]{caruana1993multitask}, low rank structure over the task parameters \citep[see e.g.][]{ando2005framework} or task clustering \citep[see e.g.][]{bakker2003task}. If a vector of ``task descriptors'' is available for each task, these can be used in MTL, see e.g.\ \citet{bonilla2007kernel}.

There is some literature on MTL for time series models. For example \citet{durichen2015multitask} consider multi-task Gaussian processes (GPs) for condition monitoring of patients, but in their case the multi-task nature is over multiple obervation channels, not patients, and they make use of GP methods for modelling the correlation between channels (see also \citealt{bonilla2008multi}).

Like us \citet{schulam2015framework} do consider multiple patients, but assume an \emph{additive} decomposition of population, subpopulation, individual and noise components, so `individual level' contributions work as offsets to the population and subpopulation effects. In contrast our eq.\ (\ref{eq:gen-model}) uses a more general low-rank structure. \citet{alaa2018personalized} use patient-specific covariates to personalise a mixture of GPs, but control inputs are not considered, and personalisation is restricted to a fixed set of subtypes. In Automatic Speech Recognition a similar concept called `i-vectors' \citep{kenny2005joint} is used for speaker personalisation, but does not adapt dynamics or handle control inputs.

A key assumption made by multi-task GP approaches is that tasks are a linear combination of the same set of underlying processes, which is usually inappropriate if the experimental units are separated in time and/or space. We avoid this difficulty by performing multi-task learning of the \emph{parameters} rather than the \emph{processes}. Our MTL framework is therefore not merely performing customisation across similar tasks (as in the independent data case), nor gaining strength over latent processes (as for MT GP models), but performing model customisation at the level of experimental units.

Our methodology is in some sense similar to \emph{random effects} used in the frequentist statistical context, which has seen some development for time series. However, the purpose of such work has largely been to gain strength over individual parameters within a cohort. We are instead looking to exploit relationships between parameters to facilitate faster adaptation and improved prediction in an MTL sense\footnote{This is an asymmetric use of MTL in the terminology of \citet{xue2007multi}.}. Furthermore,
random effects are usually applied in specific ways, such as only to the emissions \citep{tsimikas1997mixed} or only to the dynamics \citep{zhou2013nonlinear}, and with non-general learning algorithms.

\section{EXPERIMENTS}
\label{sec:experiments} 
In this section we discuss our experimental set-up and results. Section \ref{sec:experiments:setup} introduces the clinical data source, models and training objective, Section \ref{sec:experiments:results} summarises the results and Section \ref{sec:experiments:discussion} concludes with a discussion.

\begin{table*}
\small
\caption{20 and 40-step RMSE and 40-step negative log likelihood (NLL) for all channels, calculated out-of-sample. For all metrics, smaller is better.}
\vspace{5pt}
\label{tbl:BPsys-summary-mean}
\centering
\begin{tabular}{llrrrrrr}
\toprule
{} & {} & \multicolumn{3}{c}{$t=12$ m} & \multicolumn{3}{c}{$t=24$ m} \\
\cmidrule(r){3-5}\cmidrule(r){6-8}
{} &  {} & RMSE &  RMSE & NLL &  RMSE &  RMSE & NLL \\
{} &  {} & 20-ahead &  40-ahead & 20-ahead & 20-ahead & 40-ahead & 20-ahead \\
\cmidrule(r){1-2}\cmidrule(r){3-5}\cmidrule(r){6-8}
& Cohort  &      6.62 &      7.78 &  \bf 133.84 &      7.44 &      7.51 &  131.08 \\
& MTL-5   &  \bf 6.38 & \bf 7.71 &  138.59 &      5.43 &      6.15 &  124.88 \\
BPsys & MTL-7   &      6.43 &      7.77 &  136.91 &     \bf  5.42 &    \bf   6.00 &  \bf 124.05 \\
& STL     &     11.34 &     13.65 &  159.61 &      7.03 &      8.48 &  140.22 \\
& Task-ID  &      6.82 &      8.16 &  137.23 &      8.39 &      8.87 &  138.31 \\
\cmidrule(r){1-2}\cmidrule(r){3-5}\cmidrule(r){6-8}
& Cohort &     \bf  3.74 &   \bf    4.27 & \bf  110.76 &      4.31 &      4.32 & 109.36 \\
& MTL-5  &      4.08 &      4.69 & 118.23 &    \bf   3.58 &    \bf   3.88 &\bf  109.07 \\
BPdia & MTL-7  &      4.19 &      4.74 & 119.57 &      3.63 &      3.73 & 109.45 \\
& STL    &      5.46 &      6.61 & 132.20 &      5.58 &      5.91 & 126.33 \\
& Task-ID  &      4.21 &      4.92 & 117.59 &      4.36 &      4.64 & 114.07 \\
\cmidrule(r){1-2}\cmidrule(r){3-5}\cmidrule(r){6-8}
& Cohort &     10.76 &   \bf   12.28 &\bf  153.11 &     11.26 &     11.60 & 151.10 \\
& MTL-5  &     10.01 &     13.29 & 155.01 &      9.31 &     10.13 & \bf  140.74 \\
BIS & MTL-7  &    \bf  9.77  &     13.63 & 156.48 &    \bf   9.14 &   \bf    9.98 & 140.84 \\
& STL    &     14.61 &     23.25 & 184.13 &     10.06 &     12.62 & 150.43 \\
& Task-ID  &     12.35 &     14.79 & 160.21 &     12.71 &     13.27 & 155.66 \\
\bottomrule
\end{tabular}
\end{table*}

\subsection{Experimental set-up}
\label{sec:experiments:setup}

\paragraph{Data} Our data were obtained from an anaesthesia study carried out at the Golden Jubilee National Hospital in Glasgow, Scotland, as described in \citet{georgatzis2016ionlds}. These are time series of $N=40$ patients of median length 36 minutes (range approx.\ 27 - 50 mins) in length and subsampled to 15 second intervals. There are usually $d=3$ channels: BPsys, BPdia and BIS channels, but 11 patients are missing BIS. Obvious artefactual processes (such as instrument dropout or clear exogenous stimulation) were annotated and removed; all such data were marked as missing in the output data. The input data series were volumes administered by the TCI pump, which were converted into inputs $\{u_t^i\}$ by application of the PK model of \citet{white2008use} using \texttt{ode45} in MATLAB. Each patient had additional covariates of age, gender, height, weight (and BMI).

\paragraph{Models and objective}  MTL models are implemented as described in Section \ref{sec:model} and listed as MTL-$k$. The $k$ refers to the subspace dimension, excluding patient offsets, since all models infer these parameters. We also implement a task-description model `Task-D' where the $\mbf z^i$ are taken as the covariates of patients as listed above. Models are trained via maximum likelihood, and predictive accuracy is reported via Root Mean Squared Error (RMSE) and log likelihood. The variance $\tau^{-1}$ calculated during training is not optimised for prediction; we report an upper bound on likelihood by optimising $\tau$ in each predictive window, patient, channel and model. All metrics are calculated out-of-sample, for which we use a leave-one-out (LOO) prediction scheme.

In practice the iid modelling assumptions are violated significantly. This was not especially problematic during training, but caused substantial problems for inference on shorter sequences at test time. On a number of occasions, un-modelled noise processes in the data resulted in implausible predictions with high confidence. A simple approach that appears to alleviate this problem is to downsample the observation data given to the inference algorithm by a factor of 4. This removes all significant partial autocorrelations seen in the residuals; predictions are still made on the full data. A further model violation was discovered at the initial stages of the time series. Many patients had elevated vital signs (some had systolic BP above 200), which may be explained by anxiousness prior to a surgical procedure. Since we have no record of steady state vital signs, but the MTL  fit showed strong evidence of downwards bias in the first four minutes, we discarded these initial datapoints during inference at test time.

\paragraph{Benchmark models} A one-size-fits-all
\emph{cohort model} is optimised over all patients, with only the  patient-specific offsets $\{\mbf\alpha^i\}$ inferred for the predictive posterior. This is an improvement to the state-of-the-art in PK/PD modelling, which does not adapt online. A \emph{single-task model} was implemented for comparison using relatively uninformative zero-mean Gaussian priors on each parameter with standard deviation $100$. These represent the two extremes of which MTL sits in the middle.


\begin{figure*}
    \centering
    \begin{subfigure}[b]{0.8\textwidth}
        \includegraphics[trim={0 34 30 17}, clip, width=1.0\textwidth]{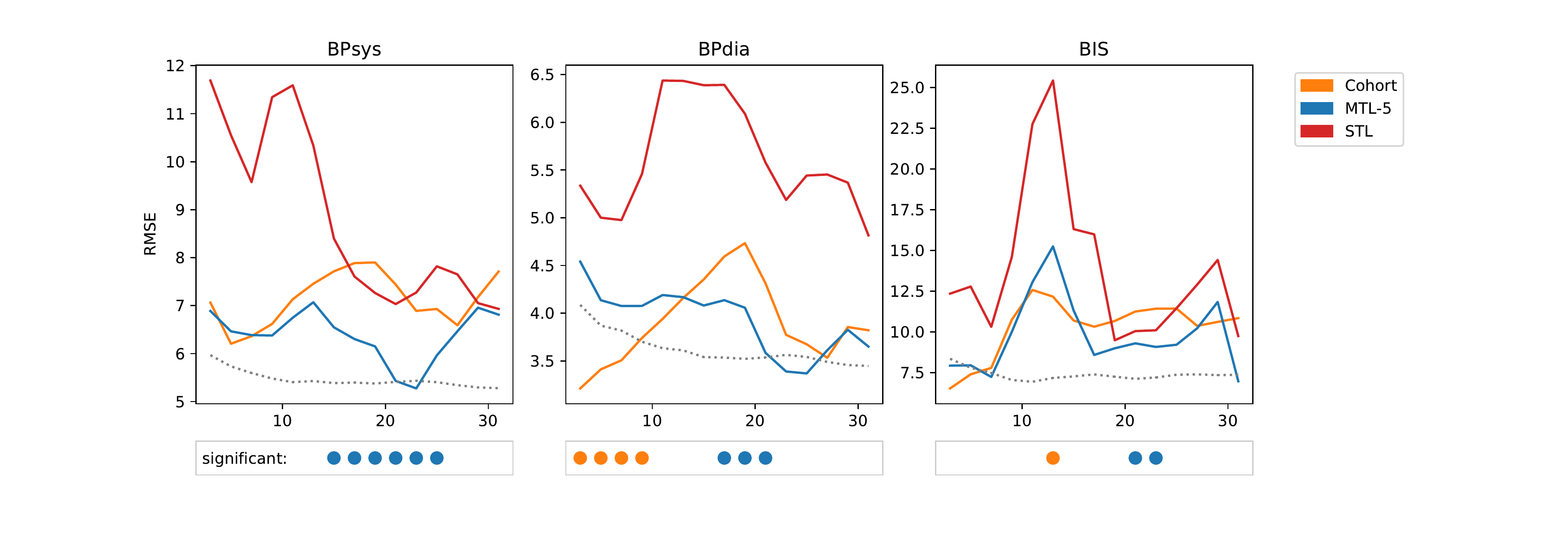}
        \caption{20-step ahead Predictive RMSE}
        \label{fig:small_t_predictive}
    \end{subfigure} 
    
    \begin{subfigure}[b]{0.8\textwidth}
        \includegraphics[trim={0 14 30 5}, clip, width=1.0\textwidth]{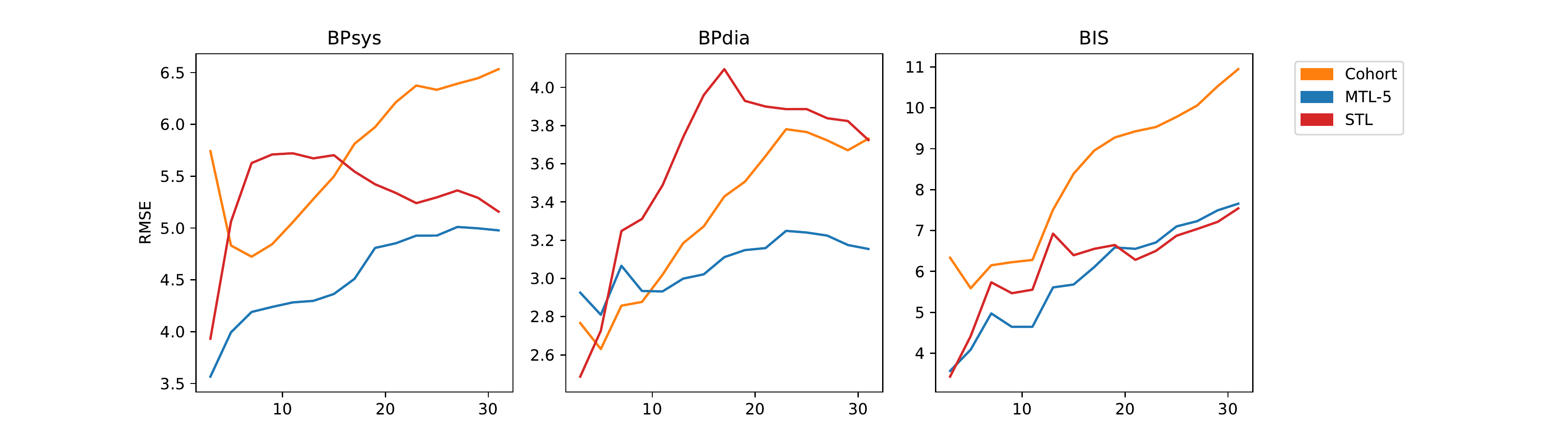}
        \caption{Retrospective RMSE (all preceding datapoints)}
        \label{fig:small_t_retro}
    \end{subfigure} 
    \caption{Performance over time (in mins). In \ref{fig:small_t_predictive}, the lower dotted line indicates the avg.\ error for the retrospective fit of the optimal PD function over each task. If the MTL or Cohort model is significantly better than the other (see text), it is shown as the relevant coloured dot below. Best viewed in colour.}
    \label{fig:small_t_performance}
\end{figure*}

\subsection{Results \label{sec:experiments:results}}
Figure \ref{fig:BIS_example} illustrates the behaviour of the MTL and cohort models on the BIS channel of one patient. The central compartment concentration $u_t$ is shown at the bottom of the plot. Credible intervals show the predictive posterior for the \emph{underlying PD function} after three different time points. The Cohort model (green) is  fixed in shape and updates its offset as more datapoints are seen; the MTL model permits much greater flexibility as seen by its adaptation and credible intervals. The MTL model is  using data from all 3 channels, but BPsys and BPdia are not shown. Only MTL-5 is shown for clarity; the performance of MTL-7 is very similar in almost all tasks. More examples are shown in supp.\ mat.

Figure \ref{fig:small_t_predictive} shows the 20-step-ahead (or 5-mins-ahead) RMSE for the Cohort, MTL and STL models for the BPsys, BPdia and BIS channels, averaged over LOO patients. Table \ref{tbl:BPsys-summary-mean} gives these results for each channel for times $t = 12$ mins and $t = 24$ mins, along with 40-step-ahead RMSE and negative log likelihood (for all columns lower is better). 
There is a clear win for MTL over STL in the plots for the first 30 minutes of infusion. The benefit is sustained across all time points in this initial period, but observe the STL model is `closing the gap' as we approach the 30 minute mark. Notice that prediction errors show some dependence on time, particularly after large changes due to infusion scheduling (see appendix \ref{supp:infusion-schedule}) at approximately 13 mins.\ and again after 27 mins.

Comparing MTL and the Cohort model, the results are closer, but MTL broadly out-performs the Cohort model on aggregate after 15 minutes of observations, although the performance drops back at the infusion change at approx.\ 27 minutes. It is not so surprising that the Cohort model does better early on, as it has fewer latent variables to estimate than the MTL model.  Many of these improvements are significant according to a $p < 0.05$ Wilcoxon signed-rank test as shown under Figure \ref{fig:small_t_predictive}, although no adjustment has been made for multiple testing (such as a Bonferroni correction). Table \ref{tbl:BPsys-summary-mean}  demonstrates that these features are also retained in a longer predictive interval.

The performance of the patient covariate model Task-D is also given in Table \ref{tbl:BPsys-summary-mean}. It performs worse in general than the Cohort model and does not perform substantially better even on the training set. This is indicative of a lack of information in our patient covariates. Use of more complex models such as Long Short Term Memory \citep[LSTM,][]{hochreiter1997long} networks in preliminary investigations also supported this conclusion. In principle one could combine the unsupervised MTL method with the task descriptors, but no benefit would be expected on this dataset.

Fig.\ \ref{fig:small_t_retro} shows the retrospective
RMSE as a function of time for the different models, based on
``post-dicting'' the the data seen up to time $t$ given the
inferences for $\mbf z$ at time $t$. These plots show that the MTL
model is substantially better at this task than the Cohort and STL models
for both BP channels.

\subsection{Discussion}
\label{sec:experiments:discussion}
The results in Section \ref{sec:experiments:results} demonstrate that the MTL framework can
achieve a predictive RMSE close to that obtained by the optimal
function in the PD class, as shown by the dotted line in Figure
\ref{fig:small_t_predictive}. This represents the predictive RMSE of
the best-fitting PD function of form (\ref{eq:PD-indiv}) fitted retrospectively per
patient on the test set.

Two major factors in the data elevate the minimum achievable RMSE for the
PD model. Firstly, even with obvious artefacts removed, there remain many
other exogenous `events' which cannot be explained by the model.
Secondly, the relative level of peaks and troughs for a given patient
(particularly in BIS) cannot always be adequately fitted with the PD model.
The changes sometimes appear more like input-driven phase changes (see e.g.
\citealt{mukamel2014transition}). Thus a more flexible (deterministic) model class may not improve the fit very much.


The performance of the Cohort PD model is surprisingly good. Traditionally, PD models have adaptive offsets based
on \emph{covariates}, and practicing anaesthetists may mentally
perform further customisation. Our work is the first instance (to our
knowledge) of investigating an online estimation of these offsets over
several channels, and our Cohort PD model appears to outperform those
generally used in practice. Nevertheless, the Cohort model does not fit well retrospectively (Fig.\ \ref{fig:small_t_retro}) which indicates both that it does not capture the inter-patient variation, and that it may perform poorly in future -- that is, if the past is representative of the future.

The worsening performance of the MTL model as $t$ approaches 30
minutes is due to a changepoint in the infusion schedule (see appendix \ref{supp:infusion-schedule}). These changepoints occur twice in the infusion sequences at
approx.\ $t = 13$ and $t = 27$ minutes and clearly present a more challenging part of the task. The data observed so far may contain little information about the response here, and perhaps in future the infusion sequence can be designed to be more informative.

An important result given by the Task-D model is the failure of patient covariates to improve performance. While there appears to be very little information about vitals \emph{shape} available in the covariates, some information about  \emph{level} may be expected had we not estimated offsets online. Note that the covariates have already been used to personalise the upstream PK model.

\section{CONCLUSION}
\label{sec:conclusion}
In this paper we have presented a method for extending the multi-task
learning framework to collections of \emph{sequential} tasks. We have
seen via use of the drug-response example that MTL can improve the
aggregate performance of a collection of discrete time input-output
dynamical systems over either a cohort or single-task
approach. Time series without control inputs can be handled equally well.
The framework is of course more general than the
use case in this paper and we are actively exploring other areas of application.

The application of this framework to the PK/PD modelling problem provides a
novel approach to personalised medicine, and at least in this case shows
substantial promise over traditional methods which personalise using patient
covariates. Where patient covariates contain such a low signal, an unsupervised
approach is essential for better performance, and a larger dataset \emph{per
se} will not help much.
Further improvements in this approach may be possible by incorporating artefact models and the optimal design of infusion sequences.

There are other important findings that may be of interest to a
clinical audience. For example, in defining a benchmark PD model, we have
proposed a PD model that is better than usually used in clinical practice,
which we have improved further using MTL. In some tasks, correlations learned
between channels via latent variables have led to useful predictions even for
a channel that has dropped out. Finally, the loading matrix $\Psi$ may be
of direct interest in itself; we leave a disentangled representation to future
work.

\section*{Acknowledgements}
The authors would like to thank Stefan Schraag, Shiona McKelvie, Mani Chandra and Nick Sutcliffe for the original study design and data collection, as well as the anonymous reviewers for helpful comments. This work was supported by The Alan Turing Institute under the EPSRC grant EP/N510129/1.
    
\appendix
\section{APPENDIX}

\subsection{Infusion Schedule}
\label{supp:infusion-schedule}
The volumes of Propofol infused during each experiment followed one of two regimes: a high-low-high dose or a low-high-low dose. 18 patients were allocated to the first group, 22 patients to the second. Despite differences in regimes and volumes, the changepoints between types of dosage happened at approximately the same time. After 27-30 minutes, the MTL models are observing a low-high or high-low transition for the first time. Prior to observing this information, it is difficult to out-perform an average effect.

\begin{figure}[h!]
    \centering
    \includegraphics[width=\linewidth]{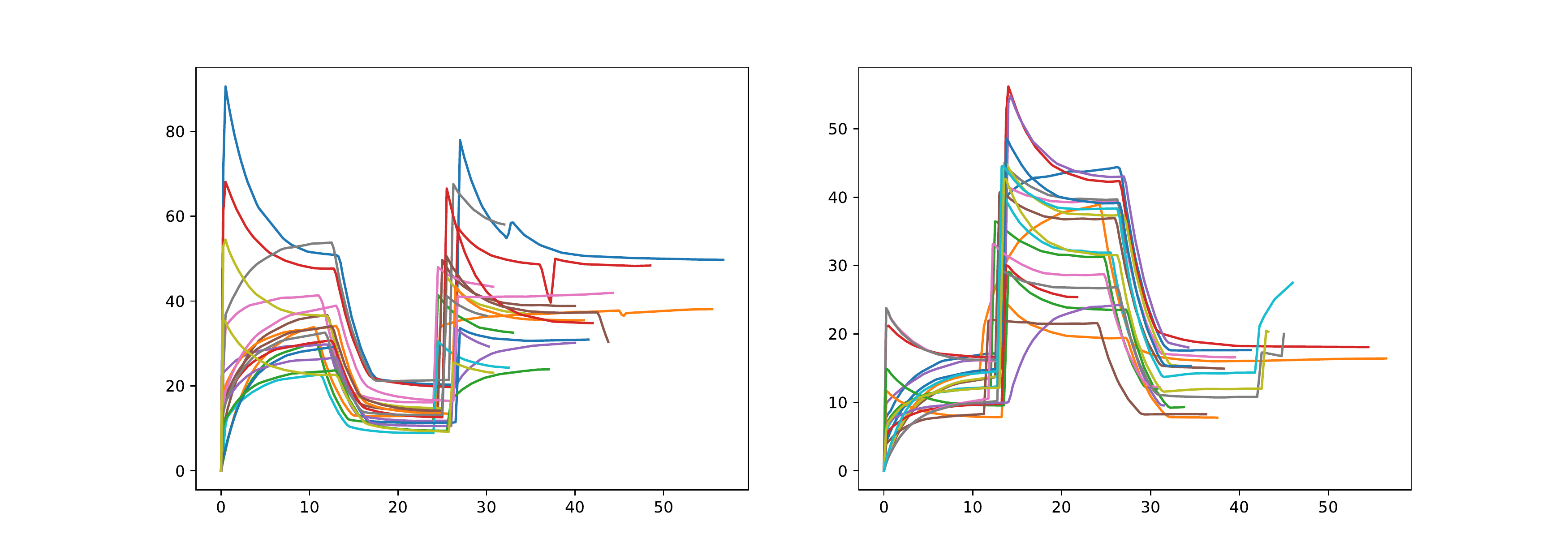}
    \caption{Inputs $\{u_t^i\}$ for all patients. Different regimes are split into separate panels for clarity.}
    \label{fig:infusion-sched}
\end{figure}

Figure \ref{fig:infusion-sched} plots each of the different inputs $\{u_t^i\}$ over all patients, split into these regimes. Note that these are the central compartment solutions to the PK equation (1) using personalised rate constants and exhibit substantial variation in magnitude.

{
\bibliography{biblio}}
\bibliographystyle{style/icml2017}


\end{document}